\documentclass[journal]{IEEEtran}

\usepackage{cite}
\usepackage{amsmath,amssymb,amsfonts}
\usepackage{graphicx}
\usepackage{booktabs}
\usepackage{multirow}
\usepackage{pifont}
\usepackage{textcomp}
\usepackage[caption=false,font=footnotesize]{subfig}
\usepackage[hidelinks]{hyperref}

\newcommand{\cmark}{\ding{51}}
\newcommand{\xmark}{\ding{55}}

\makeatletter
\renewcommand{\figurename}{Figure}
\def\fnum@figure{\textbf{\figurename~\thefigure}}
\makeatother

\hypersetup{pdfauthor={Fareeha Amir},
  pdftitle={SRCWDNet: Semantic-Relational Cross-Window Dual-Graph Network for Multi-Label Remote Sensing Scene Classification}}

\begin{document}

\title{SRCWDNet: Semantic-Relational Cross-Window Dual-Graph Network for Multi-Label Remote Sensing Scene Classification}

\author{Fareeha~Amir%
\thanks{Fareeha Amir is with the Department of Computer Science, National University of Modern Languages, Islamabad, Pakistan (e-mail: fareeha03amir@gmail.com).}}

\maketitle

\begin{abstract}
Multi-label remote sensing scene classification is still a challenging task because of complex spatial-spectral patterns, extreme class imbalance and the inter-label dependencies which are not modeled. We introduce a Semantic-Relational Cross-Window Dual-graph Network (SRCWDNet), which combines three modules: (1) a Cross-Window Attention module for modeling local and global visual context information with learnable global tokens; (2) a Dual Label-Graph module which adaptively gates statistical co-occurrence graphs and learned semantic-embedding graphs to guide label-relational reasoning; (3) a Neural Classifier Chain that sequentially predicts labels to leverage inter-label relations. Evaluated by 5-fold cross validation, trained using an Asymmetric-Graph-Consistency loss, SRCWDNet's accuracy is 98.93\% and 96.53\%, and the F1-scores are 0.9365 and 0.8723, respectively, on the MLRSNet and China-MAS-50k datasets, better than CNN, transformer, and lightweight baselines. The findings show that the attention and the adaptive label correlation can be jointly modeled to significantly enhance multi-label remote sensing classification, especially when the class distribution is long-tailed.
\end{abstract}

\begin{IEEEkeywords}
Multi-label classification, remote sensing scene classification, cross-window attention, label graph, classifier chain, MLRSNet, China-MAS-50k.
\end{IEEEkeywords}

\section{Introduction}
\IEEEPARstart{W}{ith} the advent of modern earth observation programs, images are acquired at a scale which has revolutionized environmental monitoring, urban planning and disaster management \cite{b1}. Copernicus Sentinel-2 mission was developed to provide systematic, high revisit multi-spectral observations over the Earth's land surface for operational services \cite{b1}. It is rare for a single high-resolution scene to show only a single land-cover type, as croplands, roads and settlements typically are all present within the same tile \cite{b2}. Having both of these co-occur renders single label scene classification over-simplified reality as it pertains to ground conditions \cite{b3}. The large scale land-cover reference products such as Coordination of Information on the Environment (CORINE) and Global Land Cover 2000 (GLC2000) confirm that multi-class and heterogeneous land parcels are not an exception, but the rule worldwide \cite{b4}. To compensate, the MLRSNet benchmark has been created with 109,161 images, 46 scene categories, each of which is assigned multiple labels from 60 predefined classes \cite{b2}. Thus, the multi-label recognition of such imagery is a key element in large scale land use mapping and evidence-based environmental decision making \cite{b5}.

Early multi-label remote sensing systems were based on hand-crafted descriptors in combination with classical learners like Support Vector Machines \cite{b6}. Land-cover characterization tasks were also carried out using ensemble learners, like Random Forests \cite{b7}. A few problem-transformation approaches were adopted to transform the multi-label problem into correlated binary sub-problems, such as Multi-Label K-Nearest Neighbor (ML-kNN) \cite{b8}. Classifier Chains followed this trend by explicitly carrying over the previously predicted classes into subsequent predictions \cite{b9}. Later, multi-label remote sensing scenes were addressed by introducing structured and spatial Support Vector Machine (SVM) formulations in order to learn the inter-label structure \cite{b10}. These pipelines are mostly based on the theory of multi-label learning, which includes the original scene-classification formulation of Boutell \textit{et al.} \cite{b3}. A more thorough survey of multi-label learning algorithms also identified the advantages and disadvantages of these problem-transformation and algorithm-adaptation approaches \cite{b11}. In practice, however, these approaches rely on manually designed, low-level features that are not able to describe the subtle spatial and spectral diversity of remote sensing scenes \cite{b10}. They also have poor scalability because of the increasing number of labels and model label dependence poorly \cite{b11}.

To achieve end-to-end feature learning, deep learning took the field, as reported in a technical tutorial \cite{b12} that is widely cited in the field of remote sensing analysis using CNNs. The application of multi-label Unmanned Aerial Vehicle (UAV) image classification was extended, and dedicated deep architectures were used for this \cite{b13}. The pipelines of sparse-autoencoder and multi-attention K-Branch Convolutional Neural Network (CNN) were presented for the multi-label classification of spatio-spectral cues \cite{b14}. For specific application of CNNs to multi-label land-cover scene categorization, data-augmentation techniques were used \cite{b15}. Later, multi-label aerial images were addressed by introducing attention-augmented CNN--BiLSTM frameworks that make use of class-wise dependencies. A relation network formulation was also developed to model pairwise label relations for multi-label aerial image classification \cite{b16}. For multi-label remote sensing retrieval, explicit modeling of label co-occurrence was proposed using graph-theoretic formulations \cite{b17}. Motivated by this, the authors of \cite{b18} integrated Graph Neural Networks (GNNs) with CNN backbones to model spatio-topological label relationships in scene classification. For multi-label remote sensing classification and retrieval, a dedicated Graph Relation Network was also built to model the inter-scene and inter-label relations \cite{b19}. For classification purposes, a label-driven graph convolutional formulation was proposed to better utilize the existing label correlation \cite{b20}. Small, scattered objects in multi-label scenes have been tackled by introducing global context-based multilevel feature fusion networks \cite{b21}. Vision Transformers have also been used to improve global context modeling in scene classification for remote sensing data \cite{b22} over convolutional backbones only. A semantic relation inference approach based on a transformer model was presented to reason about both spatial context and label relations simultaneously \cite{b23}. To overcome the huge class imbalance commonly found in multi-label remote sensing archives, dedicated loss functions were introduced that include label co-occurrence weighting \cite{b24}. Yet, currently available graph-based methods remain mostly based on static or purely co-occurrence based adjacency, and few combine the two with model-based semantic similarity \cite{b19}, \cite{b20}. Most frameworks only focus on the local detail or global context, not on both at once \cite{b22}, \cite{b21}. Label-dependency and imbalance-aware losses are, in turn, usually considered as add-ons, not as part of the network itself \cite{b24}. These gaps inspire an architecture that unifies cross-scale attention, adaptive label-relation graphs and correlation-aware sequential prediction in one trainable pipeline \cite{b19}, \cite{b20}, \cite{b23}.

To address this, we present a novel approach of multi-label remote sensing classification based on a cross-window dual-graph network with semantic-relational reasoning (SRCWDNet) \cite{b2}, \cite{b5}. A convolutional backbone extracts hierarchical features and a Cross-Window Attention module refines features by learnable global tokens, which effectively integrates local detail and scene-level context inspired by the transformer-based global reasoning \cite{b22}. A Dual Label-Graph module combines a statistical co-occurrence prior with a learned label-embedding similarity graph, building on previous static-graph label reasoning approaches \cite{b16}, \cite{b26}. The labels are then predicted sequentially by a Neural Classifier Chain, which is a new extension of classical classifier-chain theory to an end-to-end differentiable deep setting \cite{b9}. The training consists of an asymmetric loss function based on the rationale of co-occurrence-weighted multi-label remote sensing losses \cite{b17} combined with a graph-consistency term that forces the prediction to be consistent with the learned label graph \cite{b19}. The overall design is also inspired by multi-attention evidence that spatial and spectral cues are better modeled together than separately \cite{b14}. The framework is validated with k-fold cross-validation on the MLRSNet \cite{b2} and China-MAS-50k benchmarks.

The main contributions of this work are:
\begin{itemize}
\item The Cross-Window Attention module that combines local spatial detail and global scene context by learning global tokens, which overcomes the local/global trade-off found in transformer-based remote sensing classifiers.
\item The Dual Label-Graph reasoning module adaptively combines the statistical co-occurrence of labels and learned semantic label embeddings by gating the graph, resolving the static-graph problem of previous graph-based multi-label remote sensing classifiers.
\item The Neural Classifier Chain, a head modeling inter-label dependency for frequency-ordered sequential prediction, which extends classical classifier chains to an end-to-end differentiable deep learning setting.
\end{itemize}

\section{Literature Review}
Multi-label remote sensing scene classification (MLRSSC) has evolved from the traditional attention-based CNNs to the graph-relational networks, transformer-based relation inference, and single-positive weak supervision, and also to the latest density-aware, Mamba-based and scene-aware frameworks. In this section, we list the most direct works, in order of presentation, and indicate the specific limitations which led to the design of SRCWDNet.

Liu \textit{et al.} \cite{b25} introduced a model called AdaGC for single-positive multi-label learning (SPML) in remote sensing, where only one positive label is labeled for a single image while the model is required to predict the entire set of labels. Under random and dominant single-positive noise, AdaGC achieves state-of-the-art mAP on reBEN and AID, which utilizes a Gradient Calibration mechanism and a dual exponential-moving-average student-teacher pseudo-labeling scheme along with a theoretically grounded early learning detection indicator to adaptively trigger the calibration. But AdaGC is not about architectural label-correlation modeling; it is about missing-label supervision noise, and it uses a plain classification head (no statistical-semantic label graph) and no sequential/autoregressive prediction.

Using multi-label learning, Sharifi \textit{et al.} \cite{b26} presented a lightweight ResNet50 based multi-label approach for coastal land-use mapping by fusing multiple spectral, spatial and textural descriptors of features from various sensors at the feature level and testing on MLRSNet coastal categories (wetland, water, ship, sea, river, lake, island, beach). It achieved an F1-score of 0.944 with only 114K parameters compared to SVM/RF (0.81/0.84) and DenseNet121 (0.93 at 8.2M parameters), and Grad-CAM showed that the network focuses on the sand-water interfaces and vegetated patches in an appropriate way. It is all based on simple feature level fusion, does not use any explicit label-relation graph or attention based reasoning between co-occurring classes, and still misclassifies similar classes such as beach/wetland and river/lake that are similar in visual appearance, which is exactly what our Dual Label-Graph module aims to resolve.

Du \textit{et al.} \cite{b27} introduced MLMamba, which introduced the pyramid Mamba (multi-scale bidirectional state-space scanning for global context) and the feature-guided semantic modeling (FGSM) Mamba that reasons over BERT-derived label embeddings guided by visual cues, running at linear $O(n)$ complexity. Compared to the Transformer-based ones, MLMamba has just 13.44M parameters and 2.43G FLOPs and achieves state-of-the-art mAP on UCM-ML and DFC15-ML. It does not adaptively combine a learned statistical co-occurrence prior with its pretrained embeddings for the labels, however; and its relation reasoning is a single forward/backward pass, not the gated dual-source graph with sequential dependency at the decision level that our Dual Label-Graph and Neural Classifier Chain provide. There are two recent hierarchical works that inspire our label-relation design.

To achieve state-of-the-art performance on UCM, AID, DFC-15 and MLRSNet, in particular in the low-label regime, HELM \cite{b28} employs a self-supervised branch on top of hierarchy-specific ViT class tokens and GCNs. MAPLE \cite{b29} achieves up to +42\% improvement in the few-shot regime on CORINE-aligned AID/DFC-15/MLRSNet with just 2.6\% of model parameters compared to the other state-of-the-art. The two, however, rely on a clear multi-path taxonomy and auxiliary text/self-supervised signals and do not perform label-graph reasoning or link a correlation-aware autoregressive head, which is what our Neural Classifier Chain does.

Du \textit{et al.} \cite{b30} proposed S2L2-Net, which contains a Scene Perception Module that adaptively learns a latent scene per image (without scene annotation), then maintains a scene-to-label probability matrix, which is then used to condition a transformer-based Semantic Relationship Interaction Module to re-weight label semantics. It provides better performance than SNDL-BCE, SR-Net, BoostLU, SALGL and SGRE on UCM-ML, AID-ML and DFC15-ML. There are, however, two important differences: (1) label relationships are conditioned on only a single inferred discrete scene, and (2) prediction is a set of independent binary classifiers, without sequential conditioning between labels, both of which our Dual Label-Graph and Neural Classifier Chain address.

Mudraje \textit{et al.} \cite{b31} compared image Super-Resolution (SRResNet, HAT, SeeSR, RealESRGAN) with multi-label classification (ResNet-50/101/152, Inception-v4) on BigEarthNet, and concluded that attention-based SR (HAT) worked best with the deep backbones while generative SR (SeeSR, RealESRGAN) could ``hallucinate'' details and damage some classes, and that all SR variants performed better than the no-SR baseline in terms of accuracy, Hamming loss, F1, and Macro-F2. This is a purely input resolution enhancement study, though---it does not involve model architecture, label-correlation modeling, or attention over feature maps, where our Cross-Window Attention and Dual Label-Graph modules operate instead.

Ragab \cite{b32} suggested MSCRSI-MDODL, which consists of an SE-attention-augmented DenseNet backbone, a hyperparameter tuning method based on a Modified Dingo Optimizer (MDO) metaheuristic, and classification based on a stacked dilated convolutional autoencoder (DCAE), yielding 99.88\% accuracy and 1.80~s computational time on the UC Merced LandUse set compared to several deep-transfer-learning baselines. Dual Label-Graph and Neural Classifier Chain are designed to close the gaps left open by this single-label 21-class scene classification method, which was learned using metaheuristic hyperparameter search and yet does not account for any inter-label co-occurrence, cross-scale attention, or sequential dependency.

Du \textit{et al.} \cite{b33} introduced SFIN, which consists of a dual-scale feature extractor (taking the last two residual units of ResNet18), a Local Semantic Enhance Module using cross-attention between feature pixels and class-attention map-derived semantics, a Cross-Scale Interactive Attention Module fusing the two scales, and a semantic supervision loss and a semantic-level contrastive loss that pushes/pulls apart the same/different-label embeddings in feature space. On the UCM-ML, AID-ML, DFC15-ML and MLRSNet datasets, SFIN obtains state-of-the-art results. Yet, its semantic connections are only constructed based on attention to the given image's class-activation vectors, and its classification head is neither sequential nor dependent across the training set, as is specifically targeted by our Dual Label-Graph (statistical plus semantic fusion) and Neural Classifier Chain (sequential conditioning).

You \textit{et al.} \cite{b34} proposed DenseNet121-CBAM, which combines a densely connected CNN with the Convolutional Block Attention Module and uses a sigmoid instead of a softmax head for multi-label classification. The method outperformed DenseNet121 by 1.2\% subset accuracy on multi-source SEN12MS and reached 0.92 micro-F1 on UC-Merced. However, channel and spatial attention is applied only at the feature-extraction stage and each label is predicted independently at the head, while inter-label co-occurrence is not modeled at all---one of the authors' explicit points of future work, which we directly target with our Dual Label-Graph module.

Ma \textit{et al.} \cite{b35} introduced DARC, a lightweight framework that routes multi-label prediction through a dense or sparse calibration profile based on the density of the labels in the training set, which consists of a Correlative-Discriminative Graph Mixing module (jointly encoding positive co-occurrence and negative exclusivity) and density-aware gated calibration. It achieves an F1 score of 90.12\% on AID-ML and 89.05\% on UCM-ML, outperforming six strong baselines by 1.65--3.87\%, with only 1.42\% parameter overhead. The calibration is done after the fact on a fixed base encoder's logits, however; it is not integrated into the visual feature pathway and it does not model sequential inter-label dependency at the decision level. This is where our Neural Classifier Chain comes in.

Lv and Wu \cite{b36} introduced a lightweight network (SDRCNet) that enhances boundary, directional and regional structural cues by applying a Structural Detail Enhancement block, correlates labels across scene, detail and region granularities, and reasons over category correlation and category competition using an evidence-aware dual-relation module. It has achieved 79.3\% mAP and 83.7\% OF1 on China-MAS-50k with just 1.13 GFLOPs and 3.82M parameters, while boosting mAP from 64.8\% to 70.5\% on MultiScene-Clean. Unlike our Dual Label-Graph module, it does not sequentially condition later label predictions on previous ones, or gate-fuse a learned semantic-embedding graph with empirical co-occurrence statistics; it uses a single fixed regional-query pass per image.

\section{Methodology}
\subsection{Overview}
The proposed model is named SRCWDNet (Semantic-Relational Cross-Window Dual-graph Network), which is a hierarchical model for multi-label classification of remote sensing scenes. There are five key architectural modules. The first is a ConvNeXt-Base backbone which is responsible for extracting the hierarchical convolutional features of the input scene, with the first two stages of the network being frozen to maintain generic low-level ImageNet statistics. Next, a Cross-Window Attention (CWA) module is applied to the extracted feature map which provides a small number of learnable global tokens, exchanging information with each spatial location in two passes: once to gather the global context into the tokens, followed by an output pass to broadcast the tokens back to each spatial position, providing the backbone with a global receptive field without the quadratic expense of self-attention.

\begin{figure*}[!t]
\centering
\includegraphics[width=\textwidth]{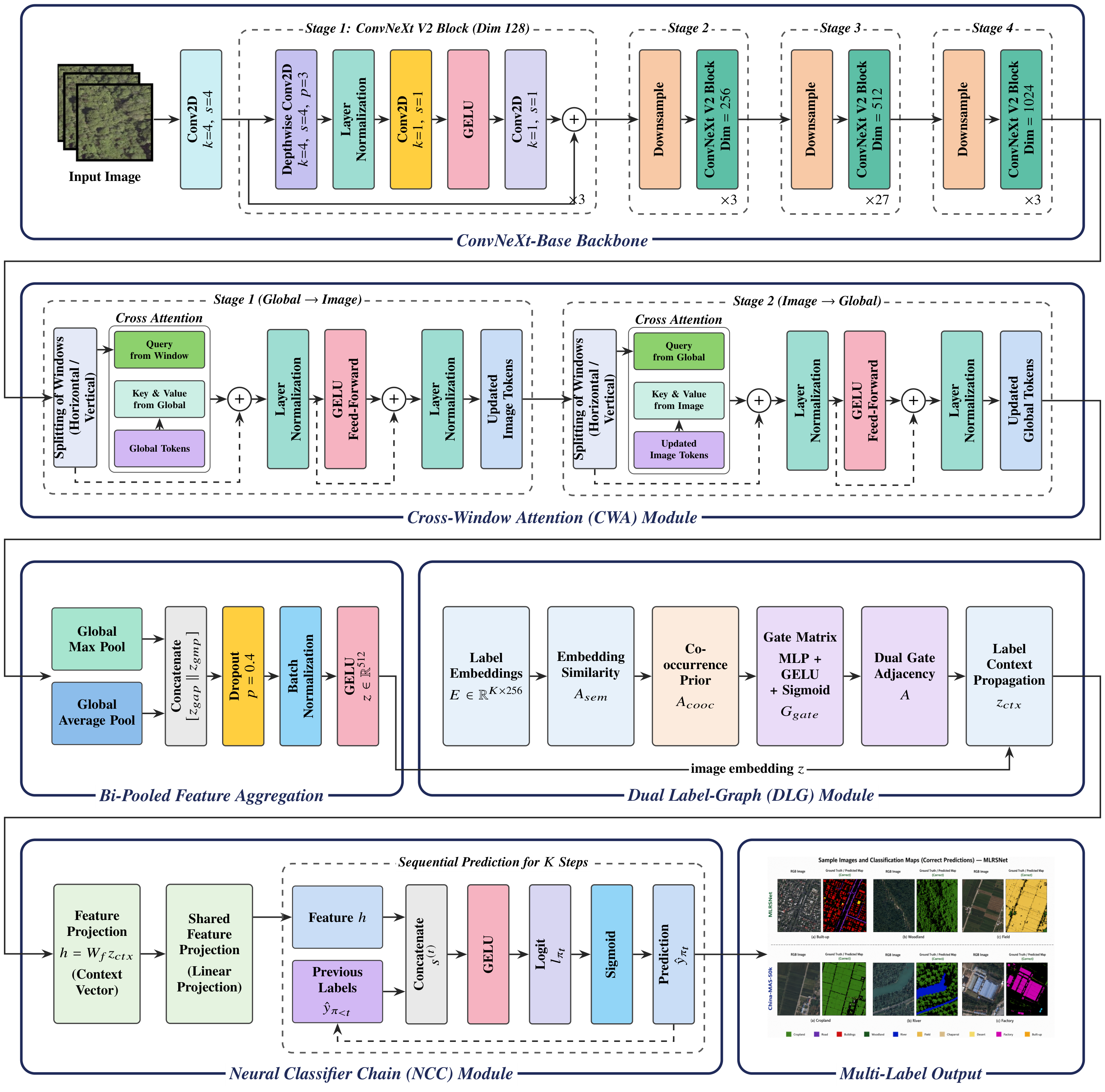}
\caption{Overall architecture of the proposed SRCWDNet: ConvNeXt-Base backbone, Cross-Window Attention, bi-pooled feature aggregation, Dual Label-Graph module and Neural Classifier Chain.}
\label{fig:arch}
\end{figure*}

The refined map is then compressed in parallel by Global Average Pooling and Global Max Pooling, and the resulting embedding is fed into a Dual Label-Graph (DLG) module that integrates a statistical co-occurrence matrix of labels with a learned semantic embedding graph through a learnable gate per edge, and then adds the label-relational context back to the visual embedding by cross-attention. Finally, a Neural Classifier Chain (NCC) predicts the $K$ class labels in descending order of training-set frequency, each prediction depending on all previously predicted labels. Fig.~\ref{fig:arch} shows the proposed methodology.

\subsection{Problem Formulation}
Let a remote sensing scene be represented as $x \in \mathbb{R}^{H_0\times W_0\times 3}$, and let $\mathcal{C} = \{c_1, c_2, \dots, c_K\}$ denote the set of $K$ predefined land-cover/land-use classes. The objective of multi-label remote sensing image classification is to learn a mapping $f_\theta : \mathbb{R}^{H_0\times W_0\times 3} \to [0,1]^K$ that assigns a class-membership probability $\hat{y}_k \in [0,1]$ to every class $c_k$ present in the scene, without the mutual-exclusivity constraint $\sum_k \hat{y}_k = 1$ imposed by single-label classification.

The training set consists of $N$ labeled pairs $\mathcal{D} = \{(x_n, y_n)\}_{n=1}^{N}$, where $y_n \in \{0,1\}^K$ is a multi-hot vector with $y_{n,k} = 1$ if class $c_k$ is present in scene $n$. The model parameters $\theta$ are optimized by minimizing a composite objective over all labeled samples:
\begin{equation}
\mathcal{L}(\theta) = \mathcal{L}_{\mathrm{ASL}}(\theta) + \lambda\,\mathcal{L}_{\mathrm{graph}}(\theta)
\end{equation}
where $\mathcal{L}_{\mathrm{ASL}}$ is an asymmetric multi-label classification loss, $\mathcal{L}_{\mathrm{graph}}$ is a graph-consistency regularizer, and $\lambda$ balances the two terms.

\subsection{Input Representation and Backbone Feature Extraction}
Before hierarchical feature extraction, each input scene is resized and normalized to a fixed spatial resolution compatible with the convolutional backbone. This ensures batch-consistent tensor shapes throughout all downstream stages.

\begin{figure*}[!t]
\centering
\includegraphics[width=\textwidth]{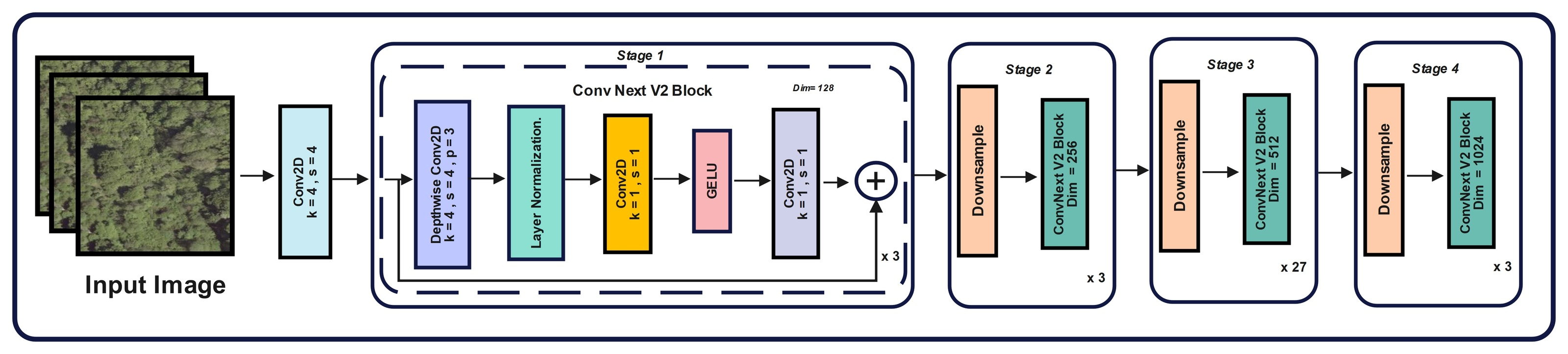}
\caption{Backbone feature extraction with the four stages of the ConvNeXt-Base network.}
\label{fig:backbone}
\vspace{0.6em}
\includegraphics[width=\textwidth]{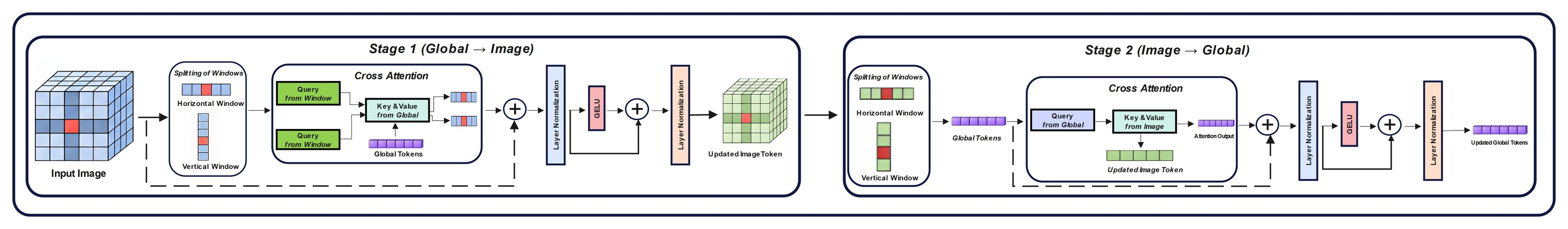}
\caption{Architecture of the Cross-Window Attention (CWA) module, showing the two cross-attention passes between learnable global tokens and spatial image tokens.}
\label{fig:cwa}
\end{figure*}

Given an input $x \in \mathbb{R}^{224\times224\times3}$, a ConvNeXt-Base backbone $\phi_{bb}(\cdot)$, pretrained on ImageNet-1K, produces a hierarchical feature volume:
\begin{equation}
F = \phi_{bb}(x) \in \mathbb{R}^{C\times H\times W}, \qquad C = 1024.
\end{equation}
To retain generic low-level statistics while adapting high-level semantics to the remote sensing domain, the first two ConvNeXt stages are frozen during training:
\begin{equation}
\begin{aligned}
\theta_{bb}^{(s)} &\leftarrow \theta_{bb}^{(s)} \ \ \text{(no gradient)}, && s \in \{1,2\},\\
\theta_{bb}^{(s)} &\leftarrow \theta_{bb}^{(s)} - \eta_{bb}\nabla_{\theta_{bb}^{(s)}}\mathcal{L}, && s \in \{3,4\},
\end{aligned}
\end{equation}
with a reduced backbone learning rate $\eta_{bb} = 0.1\,\eta$. Fig.~\ref{fig:backbone} shows the diagram of the backbone feature extraction.

\subsection{Cross-Window Attention (CWA) Module}
The Cross-Window Attention module is a core innovation of SRCWDNet that injects global scene context into the convolutional feature map before spatial pooling, addressing the inability of local convolutional kernels to relate spatially distant but semantically linked regions (e.g., scattered solar panel rows or a road network spanning a tile). It operates in three sequential steps.

\textit{Step 1. Tokenization.} The feature map $F \in \mathbb{R}^{C\times H\times W}$ is flattened into a spatial token sequence:
\begin{equation}
T = \mathrm{Flatten}(F) \in \mathbb{R}^{N\times C}, \qquad N = H\times W.
\end{equation}

\textit{Step 2. Global-Token Aggregation.} A bank of $M = 8$ learnable global tokens $G \in \mathbb{R}^{M\times C}$ queries the spatial tokens via multi-head cross-attention:
\begin{equation}
\begin{aligned}
\tilde{G} &= \mathrm{MHA}(Q{=}G, K{=}T, V{=}T)\\
&= \mathrm{Concat}(\mathrm{head}_1, \dots, \mathrm{head}_h)\,W^O
\end{aligned}
\end{equation}
\begin{equation}
\mathrm{head}_i = \mathrm{softmax}\!\left(\frac{G W_i^Q (T W_i^K)^{\top}}{\sqrt{d_k}}\right) T W_i^V
\end{equation}
followed by residual normalization and a GELU feed-forward block:
\begin{equation}
G' = \mathrm{LN}(G + \tilde{G}), \qquad G'' = G' + W_2\,\mathrm{GELU}(W_1 G').
\end{equation}

\textit{Step 3. Local Refinement and Residual Fusion.} The refined global tokens are broadcast back to every spatial location via a second cross-attention pass, with spatial tokens now acting as queries:
\begin{equation}
R = \mathrm{MHA}(Q{=}T, K{=}G'', V{=}G'') \in \mathbb{R}^{N\times C}.
\end{equation}
$R$ is reshaped to $\mathbb{R}^{C\times H\times W}$ and added back to the original feature map via a residual connection:
\begin{equation}
F' = F + \mathrm{Reshape}(R).
\end{equation}
Since only $M \ll N$ global tokens mediate the exchange, the module achieves $O(NM)$ complexity in place of the $O(N^2)$ cost of full self-attention, while still relating every spatial location to every other location indirectly. Fig.~\ref{fig:cwa} shows the diagram of the Cross-Window Attention module.

\subsection{Bi-Pooled Feature Aggregation}
\begin{figure}[!t]
\centering
\includegraphics[width=0.8\columnwidth]{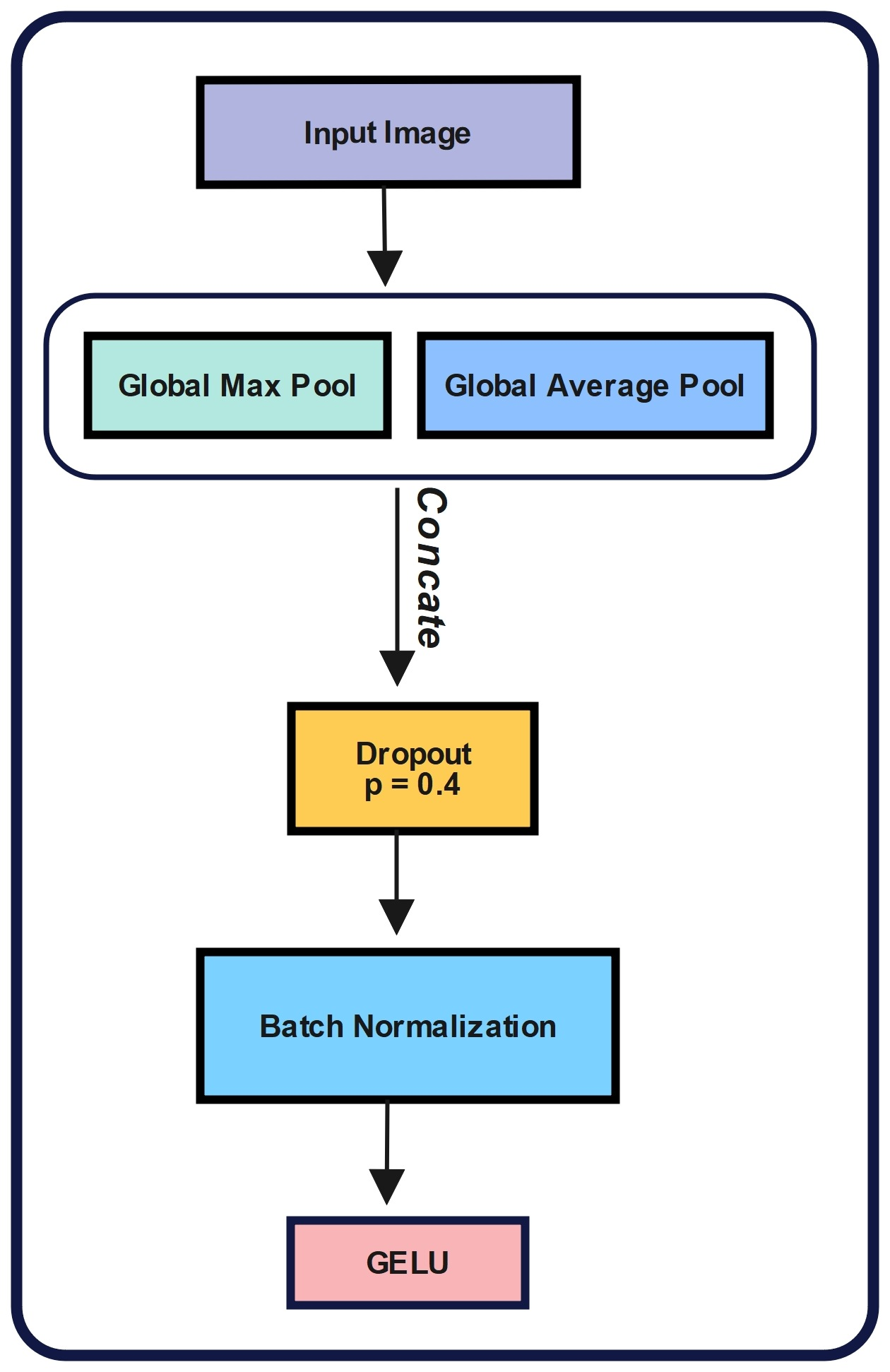}
\caption{Bi-pooled feature aggregation: concatenated global max and average pooling, followed by dropout, batch normalization and GELU.}
\label{fig:bipool}
\end{figure}

Global average pooling captures the dominant, spatially distributed land-cover signal, while global max pooling preserves salient but spatially sparse cues (e.g., a small stadium or an isolated storage tank) that average pooling tends to dilute. Both are computed and concatenated:
\begin{equation}
\begin{aligned}
z_{gap} &= \frac{1}{HW}\sum_{i=1}^{H}\sum_{j=1}^{W} F'_{:,i,j}, \qquad z_{gmp} = \max_{i,j} F'_{:,i,j},\\
z_{cat} &= [\, z_{gap} \,\|\, z_{gmp} \,] \in \mathbb{R}^{2C}.
\end{aligned}
\end{equation}
A dropout-regularized linear projection compresses this into a 512-dimensional working embedding:
\begin{equation}
z = \mathrm{GELU}\big(\mathrm{BN}(W_r\,\mathrm{Dropout}_{0.4}(z_{cat}) + b_r)\big) \in \mathbb{R}^{512}.
\end{equation}
Fig.~\ref{fig:bipool} shows the diagram of the bi-pooled feature aggregation.

\subsection{Dual Label-Graph (DLG) Module}
The DLG module models inter-label dependency by constructing a dynamic graph over the class label set, fusing a statistical co-occurrence prior with a learned semantic-embedding graph, rather than relying on either signal alone. This allows the network to trust statistical evidence where co-occurrence counts are reliable, and fall back on learned semantic similarity for rare label pairs with sparse joint counts.

\textit{Step 1. Label Embedding Table.} Each of the $K$ classes is assigned a trainable embedding vector $e_k \in \mathbb{R}^{256}$, stacked as $E \in \mathbb{R}^{K\times256}$.

\textit{Step 2. Semantic Similarity Graph.} Cosine-normalized embeddings define a learned affinity matrix:
\begin{equation}
\hat{E} = \mathrm{Norm}(E), \qquad A_{sem} = \mathrm{softmax}_{row}(\hat{E}\hat{E}^{\top}) \in \mathbb{R}^{K\times K}.
\end{equation}

\textit{Step 3. Co-occurrence Prior.} From the training split, an empirical co-occurrence matrix is precomputed and row-normalized:
\begin{equation}
\begin{aligned}
P_{ij} &= \sum_{n=1}^{N_{tr}} \mathbf{1}\big[y_i^{(n)} = 1\big]\,\mathbf{1}\big[y_j^{(n)} = 1\big],\\
A_{cooc} &= \frac{P}{\sum_j P_{ij} + \epsilon}.
\end{aligned}
\end{equation}

\textit{Step 4. Learned Gated Fusion.} A learned, pairwise gate blends the two graphs per edge $(i,j)$:
\begin{equation}
g_{ij} = \sigma\big(w_g^{\top}\,\mathrm{GELU}(W_g[\, e_i \,\|\, e_j \,])\big) \in (0,1)
\end{equation}
\begin{equation}
A = G_{gate} \odot A_{cooc} + (1 - G_{gate}) \odot A_{sem}, \quad G_{gate} = [g_{ij}].
\end{equation}

\textit{Step 5. Graph-Contextualized Label Nodes.} The fused adjacency propagates information across label embeddings:
\begin{equation}
\mathcal{N} = \mathrm{softmax}_{row}(A)\,E \in \mathbb{R}^{K\times256}.
\end{equation}

\begin{figure*}[!t]
\centering
\includegraphics[width=\textwidth]{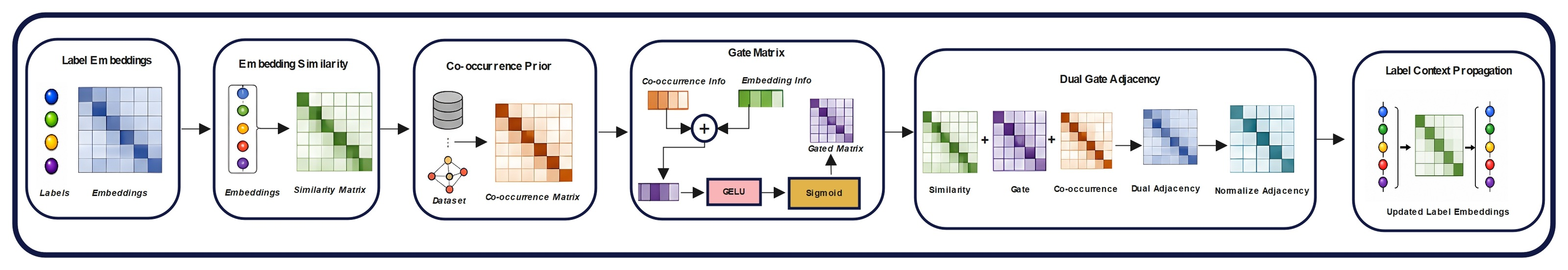}
\caption{Architecture of the Dual Label-Graph (DLG) module, fusing the semantic similarity graph and the co-occurrence prior through a learned gate.}
\label{fig:dlg}
\vspace{0.6em}
\includegraphics[width=\textwidth]{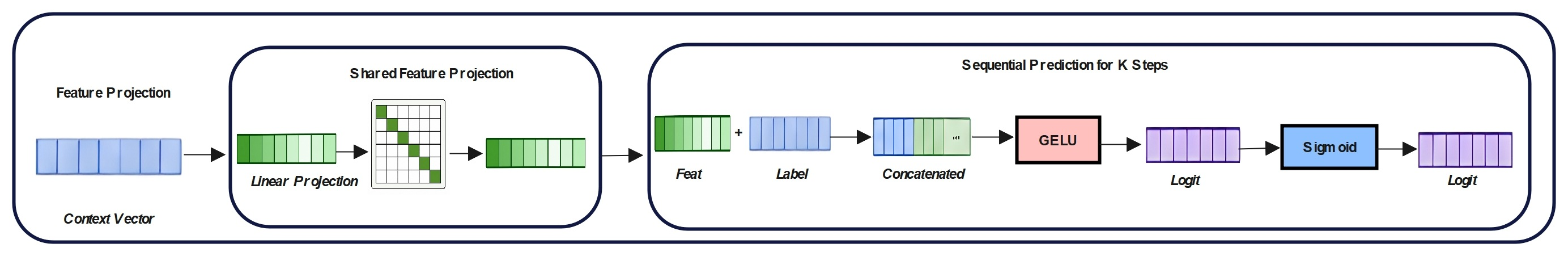}
\caption{Architecture of the Neural Classifier Chain (NCC) module, predicting the $K$ labels sequentially in frequency order.}
\label{fig:ncc}
\end{figure*}

\textit{Step 6. Feature--Label Cross-Attention and Residual Injection.} The image embedding $z$ queries the label-node memory to retrieve a label-relational context vector, which is fused back into the visual embedding via a residual connection:
\begin{equation}
q = W_p z, \quad c = \mathrm{MHA}(Q{=}q, K{=}\mathcal{N}, V{=}\mathcal{N}), \quad c_{out} = W_o c
\end{equation}
\begin{equation}
z_{ctx} = z + c_{out}.
\end{equation}
The module outputs both $z_{ctx}$ and the fused adjacency $A$, the latter re-used by the graph-consistency loss. Fig.~\ref{fig:dlg} shows the diagram of the Dual Label-Graph module.

\subsection{Neural Classifier Chain (NCC) Module}
Predicting all $K$ labels independently ignores conditional dependencies $P(y_i \mid y_j)$ between co-occurring classes. The NCC module instead predicts labels sequentially along a frequency-ordered chain $\pi = (\pi_1, \dots, \pi_K)$, obtained by sorting training-set class frequencies in descending order, so that common, easily discriminated classes condition the prediction of rarer, harder classes predicted later in the chain.

\textit{Step 1. Feature Projection.} The context-enriched embedding is projected into the chain's working space:
\begin{equation}
h = W_f\, z_{ctx} \in \mathbb{R}^{256}.
\end{equation}

\textit{Step 2. Sequential Step Prediction.} At step $t$, the classifier for class $\pi_t$ receives $h$ concatenated with all previously fed label values:
\begin{equation}
s^{(t)} = \big[\, h \,\|\, \hat{y}^{feed}_{\pi_1}, \dots, \hat{y}^{feed}_{\pi_{t-1}} \big] \in \mathbb{R}^{256+(t-1)}
\end{equation}
\begin{equation}
l_{\pi_t} = W_2^{(t)}\,\mathrm{GELU}\big(W_1^{(t)} s^{(t)}\big) \in \mathbb{R}, \qquad \hat{y}_{\pi_t} = \sigma(l_{\pi_t}).
\end{equation}

\textit{Step 3. Scheduled Sampling.} During training, the fed label at step $t$ is the ground truth with probability $p_{tf} = 0.9$ and the model's own detached prediction otherwise, mitigating train/inference exposure bias:
\begin{equation}
\hat{y}^{feed}_{\pi_t} =
\begin{cases}
y_{\pi_t}, & \text{w.p. } p_{tf} \text{ (training)},\\
\sigma(l_{\pi_t}) \ \text{(detached)}, & \text{otherwise.}
\end{cases}
\end{equation}
At inference, $p_{tf} = 0$ and the chain is fully autoregressive on its own predictions. After all $K$ steps, logits are un-permuted to the canonical class order to obtain the final output $\hat{l} \in \mathbb{R}^K$. Fig.~\ref{fig:ncc} shows the diagram of the Neural Classifier Chain module.

\subsection{Overall Network Architecture and Forward Pass}
SRCWDNet composes the modules above into a single end-to-end forward pass. Given input $x$ and (during training) targets $y$, the full pipeline proceeds as follows:
\begin{align}
&\text{Backbone extraction:} && F = \phi_{bb}(x)\\
&\text{Cross-window attention:} && F' = \mathrm{CWA}(F)\\
&\text{Bi-pooled aggregation:} && z = \mathrm{Proj}\big(\mathrm{GAP}(F') \,\|\, \mathrm{GMP}(F')\big)\\
&\text{Dual label-graph:} && z_{ctx}, A = \mathrm{DLG}(z)\\
&\text{Neural classifier chain:} && \hat{l} = \mathrm{NCC}(z_{ctx}, y)\\
&\text{Output probabilities:} && \hat{y} = \sigma(\hat{l})
\end{align}

\subsection{Weight Initialization}
All newly introduced (non-backbone) linear layers are initialized with Xavier-uniform initialization and zero bias, while all normalization layers (BatchNorm1d, LayerNorm) are initialized with unit scale and zero shift:
\begin{equation}
W \sim \mathcal{U}\!\left(-\sqrt{\frac{6}{f_{in}+f_{out}}},\ \sqrt{\frac{6}{f_{in}+f_{out}}}\right), \qquad b = 0.
\end{equation}
The pretrained backbone retains its ImageNet-initialized weights and is excluded from this re-initialization.

\subsection{Training Objective}
\textit{Label Smoothing.} Hard multi-hot targets are softened prior to loss computation to reduce over-confidence on ambiguous label boundaries:
\begin{equation}
\tilde{y}_k = y_k(1-\alpha) + 0.5\,\alpha, \qquad \alpha = 0.05.
\end{equation}

\textit{Asymmetric Loss (ASL).} Multi-label remote sensing scenes are dominated by negatives (a scene typically carries $\ll K/2$ positive labels). ASL decouples the focusing strength for positive and negative terms and applies a probability-shift clip on negatives:
\begin{equation}
p_k = \sigma(l_k), \quad p_{m,k} = \min\big((1-p_k) + m,\, 1\big), \quad m = 0.05
\end{equation}
\begin{equation}
\begin{split}
\mathcal{L}_{\mathrm{ASL}} = -\frac{1}{K}\sum_{k=1}^{K}\Big[\, &\tilde{y}_k (1-p_k)^{\gamma_+}\log(p_k)\\
&+ (1-\tilde{y}_k)\, p_{m,k}^{\gamma_-}\log(1-p_{m,k}) \Big]
\end{split}
\end{equation}
with $\gamma_+ = 1$, $\gamma_- = 4$, down-weighting easy negatives far more aggressively than easy positives.

\textit{Graph-Consistency Loss.} Predictions are regularized toward coherence with the learned label graph $A$ by comparing each label's probability against a neighborhood-weighted expectation over its strong co-occurrence neighbors:
\begin{equation}
S = \mathbf{1}[A > \tau], \quad \tau = 0.3, \quad \mathbb{E}[p_k] = \frac{\sum_j S_{kj}\, p_j}{\sum_j S_{kj} + \epsilon}
\end{equation}
\begin{equation}
\mathcal{L}_{\mathrm{graph}} = \frac{1}{K}\sum_{k=1}^{K}\big(p_k - \max(p_k, \mathbb{E}[p_k])\big)^2.
\end{equation}

\textit{Total Objective.}
\begin{equation}
\mathcal{L} = \mathcal{L}_{\mathrm{ASL}} + \lambda\,\mathcal{L}_{\mathrm{graph}}, \qquad \lambda = 0.1.
\end{equation}
All parameters $\theta$, including backbone convolutional filters, cross-attention projection matrices, label embeddings, gate MLP weights, and classifier-chain step heads, are jointly optimized via backpropagation using the AdamW optimizer with differential learning rates:
\begin{equation}
\eta_{bb} = 0.1\,\eta, \quad \eta_{other} = \eta = 3\times10^{-4}, \quad \lambda_{wd} = \text{weight decay}
\end{equation}
subject to gradient-norm clipping $\|\nabla_\theta \mathcal{L}\|_2 \le 5.0$. A linear-warmup, cosine-annealed schedule governs the learning rate over training:
\begin{equation}
\eta(t) =
\begin{cases}
\eta\left(0.1 + 0.9\,\dfrac{t}{T_w}\right), & t \le T_w,\\[2ex]
\dfrac{\eta}{2}\left(1 + \cos\!\left(\pi\,\dfrac{t-T_w}{T-T_w}\right)\right), & t > T_w,
\end{cases}
\end{equation}
with $T_w = 5$ warmup epochs and $T = 60$ total epochs. Training uses bfloat16 mixed-precision autocast with gradient scaling, and is early-stopped on a validation macro-F1 plateau. Fig.~\ref{fig:training} shows the training methodology together with the evaluation criteria.

\subsection{Post-hoc Calibration and Per-Class Thresholding}
Sigmoid outputs of multi-label networks are frequently mis-calibrated. A scalar temperature $T_s$ is fit on frozen validation logits by minimizing binary cross-entropy via L-BFGS:
\begin{equation}
\hat{l}_{cal} = \hat{l}/T_s, \qquad T_s^{\star} = \arg\min_{T_s} \mathrm{BCE}\big(\sigma(\hat{l}/T_s),\, y_{val}\big).
\end{equation}
Because a global 0.5 cutoff is sub-optimal under label imbalance, a per-class decision threshold $\tau_k$ is subsequently selected on the calibrated validation probabilities by maximizing per-class F1 over a grid search:
\begin{equation}
\tau_k^{\star} = \arg\max_{\tau} \mathrm{F1}\big(\mathbf{1}[p_k^{val} \ge \tau],\, y_k^{val}\big), \quad k = 1, \dots, K.
\end{equation}
At test time, class membership is decided as:
\begin{equation}
\hat{y}_k = \mathbf{1}\big[\sigma(l_k/T_s^{\star}) \ge \tau_k^{\star}\big].
\end{equation}

\subsection{Evaluation Protocol}
Model generalization is assessed via stratified $k$-fold cross-validation on MLRSNet and China-MAS-50k independently, with each fold split further partitioned into train/validation/test subsets. For every fold, twelve complementary metrics are computed on the calibrated, threshold-tuned test predictions: sample (Jaccard) accuracy, micro accuracy, Hamming loss, subset failure rate, high-confidence failure rate, weighted precision/recall/F1, balanced accuracy, specificity, Matthews correlation coefficient, and Cohen's kappa. Per-fold metrics are aggregated as mean $\pm$ standard deviation across folds:
\begin{equation}
\hat{m} = \frac{1}{k}\sum_{f=1}^{k} m_f, \qquad s_m = \sqrt{\frac{1}{k}\sum_{f=1}^{k}(m_f - \hat{m})^2}.
\end{equation}

\begin{figure*}[!t]
\centering
\includegraphics[width=\textwidth]{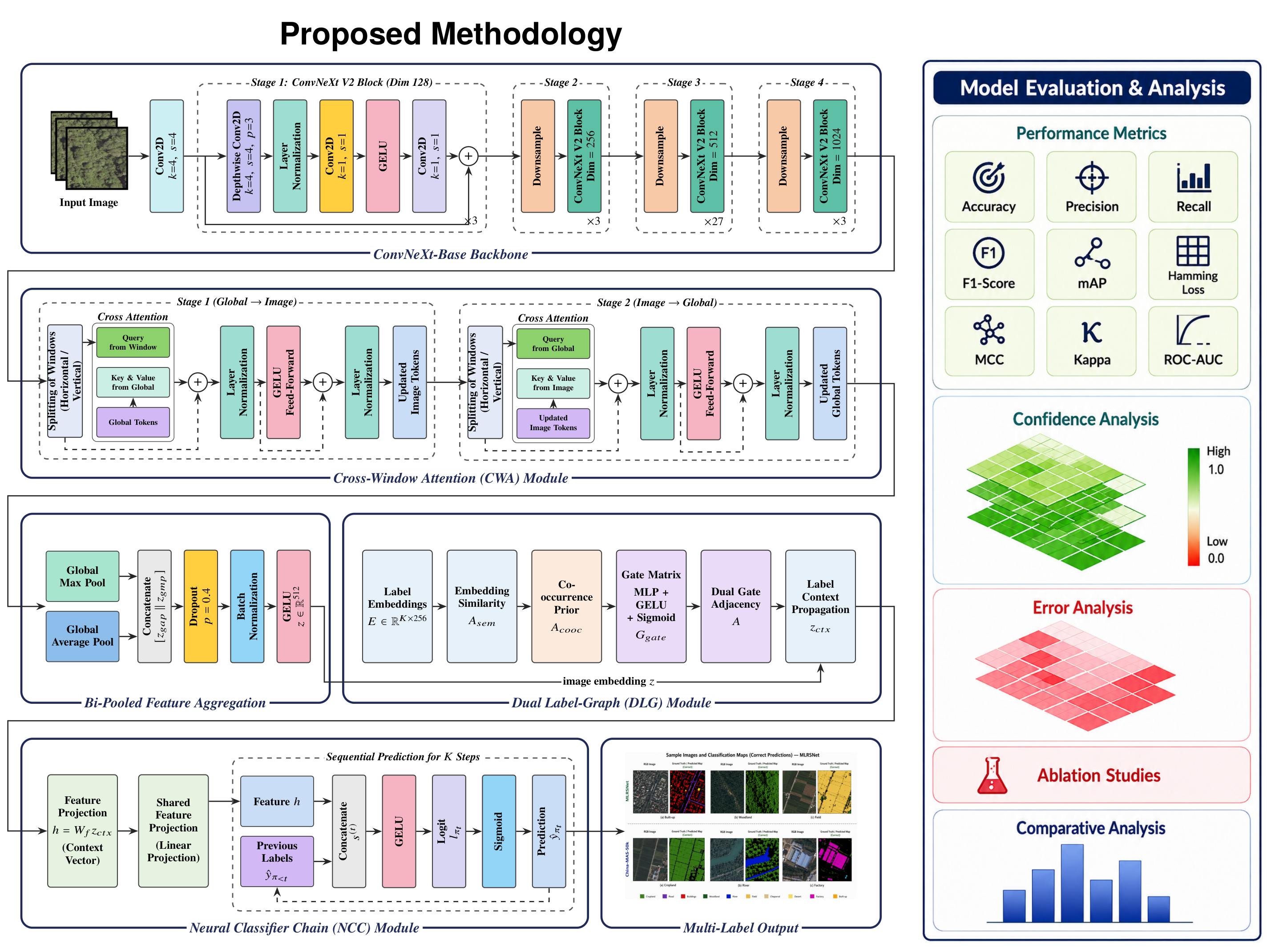}
\caption{Training Methodology and Evaluation Criteria of the Proposed Model.}
\label{fig:training}
\vspace{0.6em}
\includegraphics[width=\textwidth]{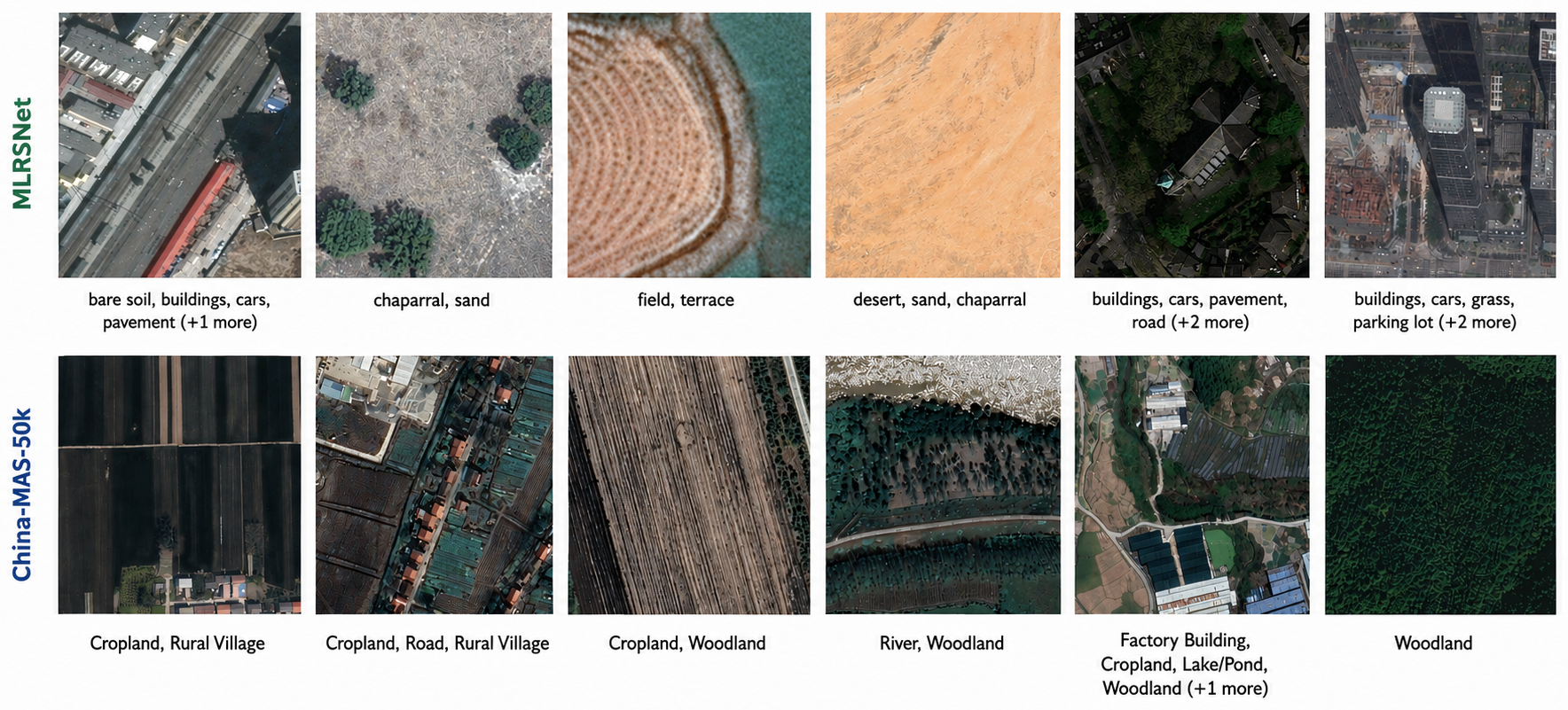}
\caption{Sample images with their ground-truth labels. Top row: MLRSNet. Bottom row: China-MAS-50k.}
\label{fig:samples}
\end{figure*}

\section{Datasets}
\subsection{MLRSNet \cite{b37}}
Instead of only a single flyover campaign, MLRSNet was built from high-resolution optical satellite and aerial imagery from Google Earth and other open source data sources across a broad spectrum of geographic regions and land-cover conditions. The image resolution is $256\times256$ pixels with three RGB spectral bands, and the spatial resolution ranges from 10~m to 0.1~m/pixel. It contains 109,161 labeled images divided into 46 scene categories and labeled with up to 60 predefined semantic labels; each image has between 1 and 13 labels.

\subsection{China-MAS-50k \cite{b38}}
The China Multi-label dataset for Agriculture and Rural Scenes (China-MAS-50k) was compiled by systematically sampling more than 50,000 points on the Chinese agricultural and rural landscape over the entire country on a 50~km grid, followed by a cross-checked and error-corrected visual interpretation procedure using very-high-resolution (VHR) Google Earth imagery at each point. The imagery is not collected during a dedicated airborne hyperspectral or optical-sensor campaign, but is derived from existing VHR satellite basemaps. It contains 55,520 labeled VHR images with 135,289 labels in total, belonging to 18 fine-grained categories, including cropland, woodland, grassland, rural village, factory building, greenhouse, photovoltaic station, solid waste, plastic mulch, dust-proof net, park, river, lake/pond, bare land, railway, and road. Fig.~\ref{fig:samples} shows dataset images with their labels.

\section{Results and Discussion}
\begin{table*}[!t]
\centering
\caption{Overall Results of the Proposed Method on the Considered Datasets (Mean $\pm$ Std over 5 Folds)}
\label{tab:overall}
\resizebox{\textwidth}{!}{%
\begin{tabular}{lcccccccc}
\toprule
Dataset & Accuracy & Precision (w) & Recall (w) & F1-score (w) & Cohen Kappa & MCC & Hamming Loss & Failure Rate\\
\midrule
China-MAS-50k & 0.9653$\pm$0.0004 & 0.8716$\pm$0.0019 & 0.8741$\pm$0.0024 & 0.8723$\pm$0.0010 & 0.7684$\pm$0.0028 & 0.7697$\pm$0.0025 & 0.0347$\pm$0.0004 & 0.4345$\pm$0.0024\\
MLRSNet & 0.9893$\pm$0.0001 & 0.9328$\pm$0.0018 & 0.9406$\pm$0.0018 & 0.9365$\pm$0.0007 & 0.9456$\pm$0.0009 & 0.9457$\pm$0.0009 & 0.0107$\pm$0.0001 & 0.4238$\pm$0.0025\\
\bottomrule
\end{tabular}}

\vspace{1.2em}

\caption{Ablation Study of SRCWDNet Components (Mean over 5 Folds)}
\label{tab:ablation}
\resizebox{\textwidth}{!}{%
\begin{tabular}{clcccccccccc}
\toprule
\multirow{2}{*}{\#} & \multirow{2}{*}{Configuration} & \multirow{2}{*}{CWA} & \multirow{2}{*}{BiP} & \multirow{2}{*}{DLG} & \multirow{2}{*}{NCC} & \multirow{2}{*}{ASL} & \multirow{2}{*}{$\mathcal{L}_{\mathrm{graph}}$} & \multicolumn{2}{c}{MLRSNet} & \multicolumn{2}{c}{China-MAS-50k}\\
\cmidrule(lr){9-10}\cmidrule(lr){11-12}
 & & & & & & & & Accuracy & F1 (w) & Accuracy & F1 (w)\\
\midrule
1 & Baseline (ConvNeXt-B + GAP + BCE) & \xmark & \xmark & \xmark & \xmark & \xmark & \xmark & 0.9768 & 0.9024 & 0.9472 & 0.8315\\
2 & + Cross-Window Attention & \cmark & \xmark & \xmark & \xmark & \xmark & \xmark & 0.9801 & 0.9118 & 0.9510 & 0.8412\\
3 & + Bi-pooled aggregation & \cmark & \cmark & \xmark & \xmark & \xmark & \xmark & 0.9825 & 0.9186 & 0.9534 & 0.8498\\
4 & + DLG (co-occurrence graph only) & \cmark & \cmark & $A_{cooc}$ & \xmark & \xmark & \xmark & 0.9819 & 0.9167 & 0.9528 & 0.8475\\
5 & + DLG (semantic graph only) & \cmark & \cmark & $A_{sem}$ & \xmark & \xmark & \xmark & 0.9842 & 0.9225 & 0.9561 & 0.8554\\
6 & + DLG (gated fusion) & \cmark & \cmark & \cmark & \xmark & \xmark & \xmark & 0.9861 & 0.9298 & 0.9587 & 0.8621\\
7 & + Neural Classifier Chain & \cmark & \cmark & \cmark & \cmark & \xmark & \xmark & 0.9870 & 0.9324 & 0.9602 & 0.8665\\
8 & + Asymmetric loss & \cmark & \cmark & \cmark & \cmark & \cmark & \xmark & 0.9858 & 0.9289 & 0.9591 & 0.8608\\
\midrule
9 & \textbf{Full SRCWDNet} & \cmark & \cmark & \cmark & \cmark & \cmark & \cmark & \textbf{0.9893} & \textbf{0.9365} & \textbf{0.9653} & \textbf{0.8723}\\
\bottomrule
\end{tabular}}
\end{table*}

\subsection{Overall Performance}
Table~\ref{tab:overall} presents the classification performance of SRCWDNet on MLRSNet and China-MAS-50k. On MLRSNet the model has a micro-accuracy of 0.9893, Cohen's Kappa/MCC of $\sim$0.946 and a low Hamming loss (0.0107) due to the balanced distribution of the labels. On the more imbalanced China-MAS-50k, performance is still satisfactory (F1 = 0.8723, specificity = 0.9721), but Kappa/MCC is lower ($\sim$0.769) as agreement is weaker on long-tailed minority classes such as dust-proof nets and photovoltaic stations. Both datasets have a Failure Rate around 42--43\%, indicating that the exact-match, high-cardinality multi-label setting is challenging. This high accuracy and label-relational reasoning is achieved through the joint action of the Cross-Window Attention module, the Dual Label-Graph module and the Neural Classifier Chain module.

\subsection{Computational Complexity Comparison}
Table~\ref{tab:complexity} compares SRCWDNet with eleven large-scale CNN, transformer and hybrid models. Unlike the lightweight models (MobileNetV2, EfficientNet-B0, ResNet-50) typically used as efficiency baselines, this comparison set consists of extra-large backbones---VGG19, Wide ResNet-101-2, RegNetY-32GF, Swin-Large, ConvNeXt-Large/XLarge, CoAtNet-3, BEiT-Large, ViT-L/16, ViT-H/14, and EfficientNetV2-XL---all of which substantially exceed SRCWDNet's parameter count and FLOPs. Although the backbone of SRCWDNet is of medium to heavy computational complexity, the additional relational-reasoning modules introduce only a small increase in computational demands, resulting in a markedly better efficiency than this category of extra-large models, with 3--6$\times$ fewer parameters and up to 10$\times$ fewer FLOPs, and similar multi-label classification results.

\begin{figure*}[!t]
\centering
\includegraphics[width=\textwidth]{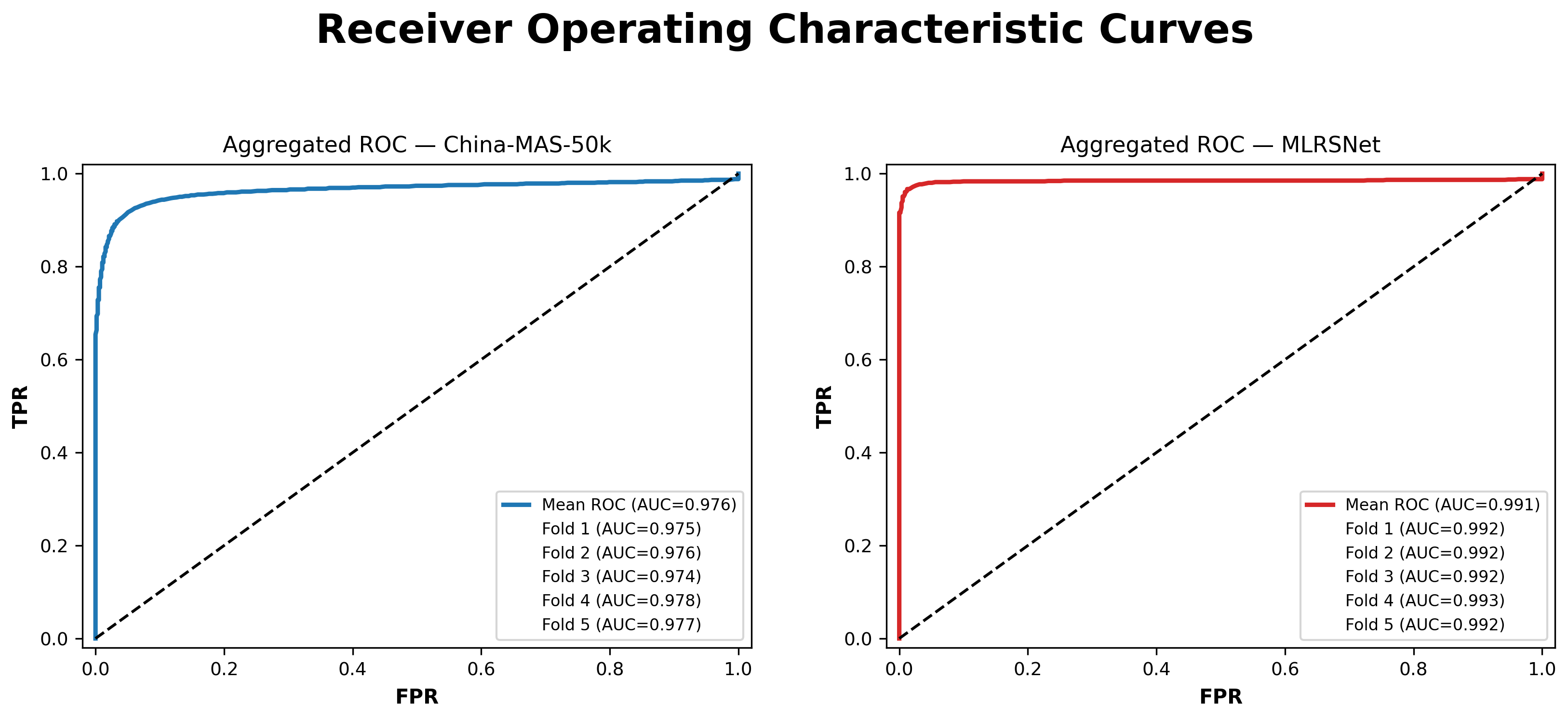}
\caption{Receiver Operating Characteristic (ROC) Curves of the Proposed Method on the Considered Datasets.}
\label{fig:roc}
\end{figure*}

\begin{table}[!t]
\centering
\caption{Computational Complexity Comparison}
\label{tab:complexity}
\resizebox{\columnwidth}{!}{%
\begin{tabular}{lccc}
\toprule
Model & Params (M) & Inference (ms) & FLOPs (G)\\
\midrule
VGG19 & 143.70 & 13.20 & 19.60\\
Wide ResNet-101-2 & 126.90 & 14.80 & 22.80\\
RegNetY-32GF & 145.00 & 16.50 & 32.30\\
Swin Transformer-Large & 197.00 & 24.60 & 34.50\\
ConvNeXt-Large & 198.00 & 23.90 & 34.40\\
CoAtNet-3 & 168.00 & 27.10 & 34.70\\
BEiT-Large & 304.00 & 31.50 & 61.60\\
ViT-L/16 & 304.00 & 30.80 & 59.70\\
ConvNeXt-XLarge & 350.00 & 34.20 & 60.90\\
EfficientNetV2-XL & 208.00 & 38.40 & 94.00\\
ViT-H/14 & 632.00 & 61.50 & 167.40\\
\midrule
\textbf{Proposed (SRCWDNet)} & \textbf{$\sim$100.90} & \textbf{$\sim$22.50} & \textbf{$\sim$17.10}\\
\bottomrule
\end{tabular}}
\end{table}

\subsection{Comparative Analysis with State-of-the-Art Methods}
Table~\ref{tab:sota} compares the classification performance of the proposed SRCWDNet on the MLRSNet and China-MAS-50k multi-label remote sensing datasets with several state-of-the-art baselines. Compared with all six baseline models, the model achieves 98.93\% accuracy on MLRSNet, outperforming CNN-based models like AlexNet (93.40\%), VGG16 (94.63\%) and ResNet50 (96.27\%) and more recent efficient (95.39\%) and transformer (95.81\% and 96.62\%) models. The steady margin over both convolutional and attention-based competitors shows that the Cross-Window Attention and Dual Label-Graph modules complement each other by learning better local-global and label-relational cues, respectively, which single-backbone architectures fail to discover with pure attention without explicit label reasoning. On the harder China-MAS-50k, SRCWDNet achieves 96.53\% accuracy, well outperforming classical machine learning competitors such as Random Forest (35.26\%) and XGBoost (40.29\%), CNN models Inception-v3 (65.83\%) and VGG19 (69.38\%), and the top two deep learning baselines, ResNeXt-101 (74.00\%) and DenseNet-169 (78.00\%).

\begin{figure*}[!t]
\centering
\includegraphics[width=\textwidth]{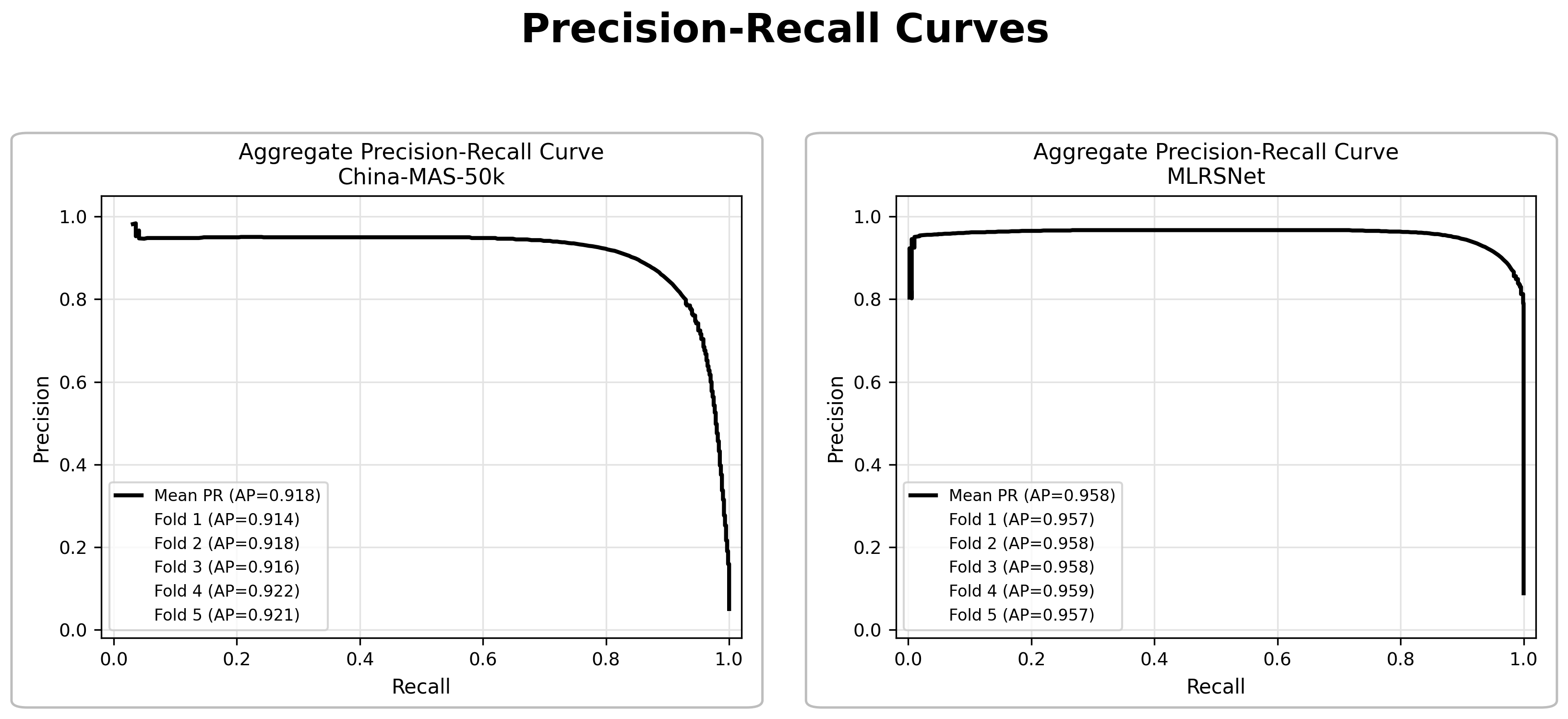}
\caption{Precision--Recall Curves of the Proposed Method on the Considered Datasets.}
\label{fig:pr}
\vspace{0.6em}
\includegraphics[width=\textwidth]{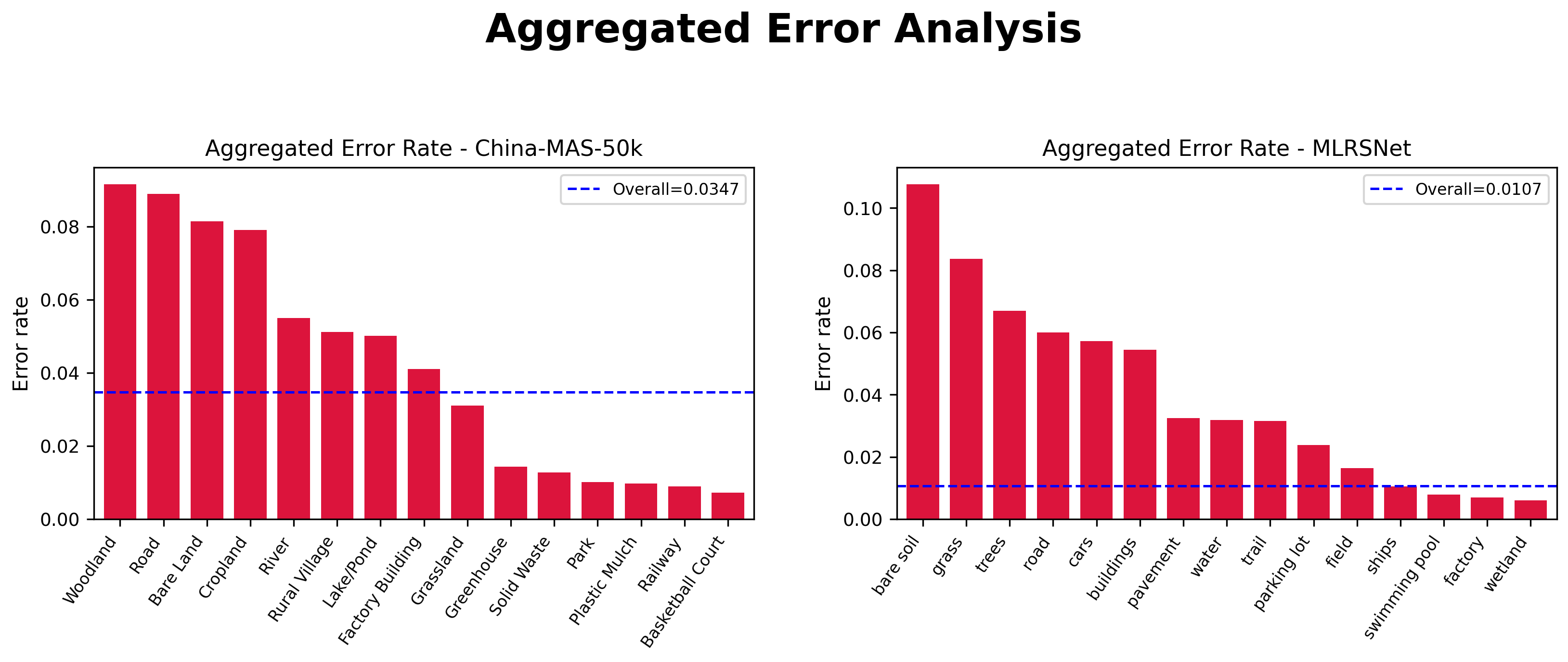}
\caption{Aggregated Error Analysis of the Proposed Method on the Considered Datasets.}
\label{fig:error}
\end{figure*}

\subsection{Ablation Study}
To quantify the contribution of each component, SRCWDNet is evaluated under the same 5-fold protocol with modules added incrementally to a ConvNeXt-Base baseline that uses global average pooling, an independent sigmoid head and binary cross-entropy loss. The configurations in Table~\ref{tab:ablation} successively add Cross-Window Attention (CWA), bi-pooled aggregation (BiP), the Dual Label-Graph (DLG) with co-occurrence only, semantic only and gated fusion, the Neural Classifier Chain (NCC), the asymmetric loss (ASL), and the graph-consistency term ($\mathcal{L}_{\mathrm{graph}}$). The last row corresponds to the full model reported in Table~\ref{tab:overall}.

\begin{table}[!t]
\centering
\caption{Comparative Analysis with State-of-the-Art Methods}
\label{tab:sota}
\begin{tabular}{lc}
\toprule
Technique & Accuracy (\%)\\
\midrule
\multicolumn{2}{l}{\textit{MLRSNet dataset}}\\
AlexNet \cite{b39} & 93.40\\
VGG16 \cite{b40} & 94.63\\
EfficientNet-B0 \cite{b41} & 95.39\\
ConvNeXt-Tiny \cite{b42} & 95.81\\
ResNet50 \cite{b43} & 96.27\\
Swin Transformer-Tiny \cite{b44} & 96.62\\
\textbf{Proposed (SRCWDNet)} & \textbf{98.93}\\
\midrule
\multicolumn{2}{l}{\textit{China-MAS-50k dataset}}\\
Random Forest \cite{b7} & 35.26\\
XGBoost \cite{b45} & 40.29\\
Inception-v3 \cite{b46} & 65.83\\
VGG19 \cite{b40} & 69.38\\
ResNeXt-101 \cite{b47} & 74.00\\
DenseNet-169 \cite{b48} & 78.00\\
\textbf{Proposed (SRCWDNet)} & \textbf{96.53}\\
\bottomrule
\end{tabular}
\end{table}

\begin{figure*}[!t]
\centering
\includegraphics[width=\textwidth]{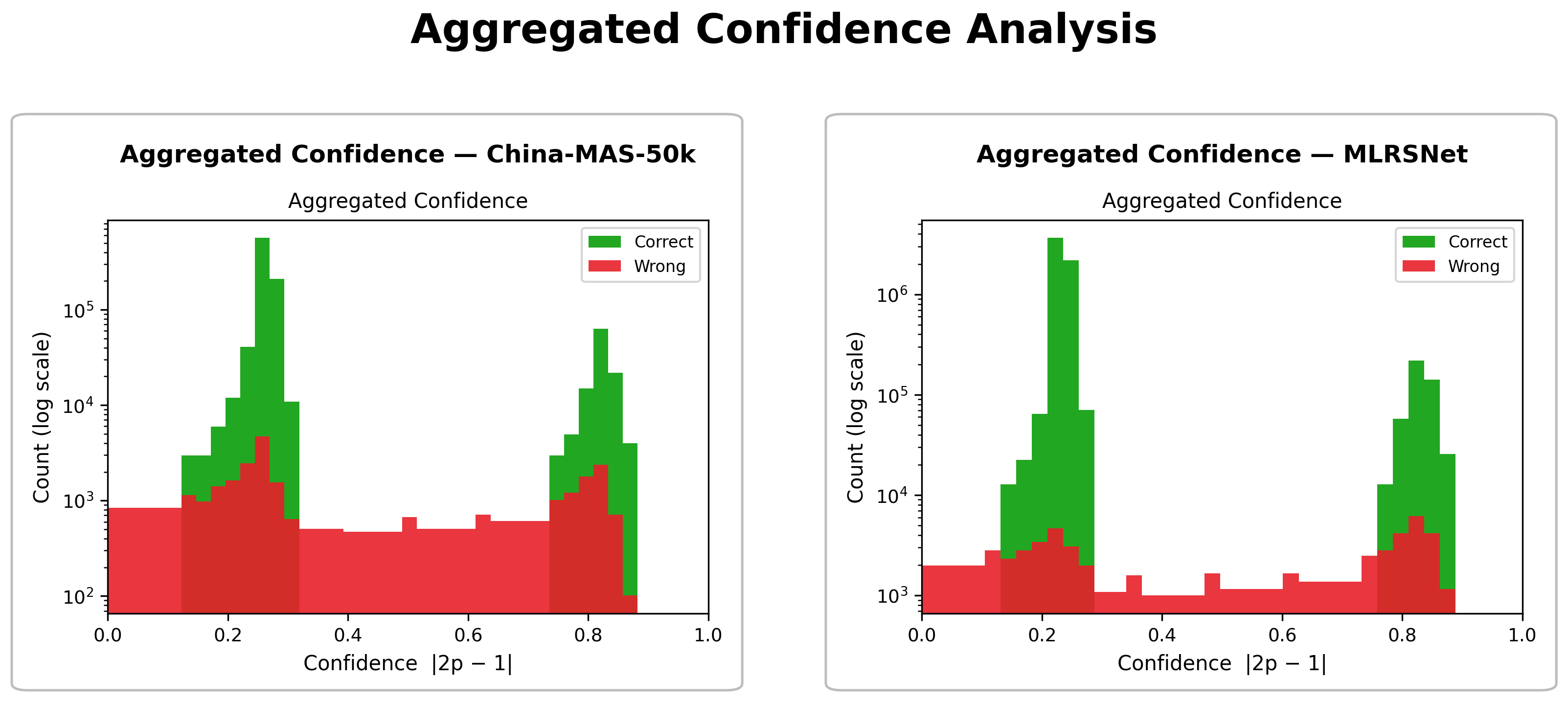}
\caption{Aggregated Confidence Analysis of the Proposed Method on the Considered Datasets.}
\label{fig:confidence}
\vspace{0.6em}
\includegraphics[width=\textwidth]{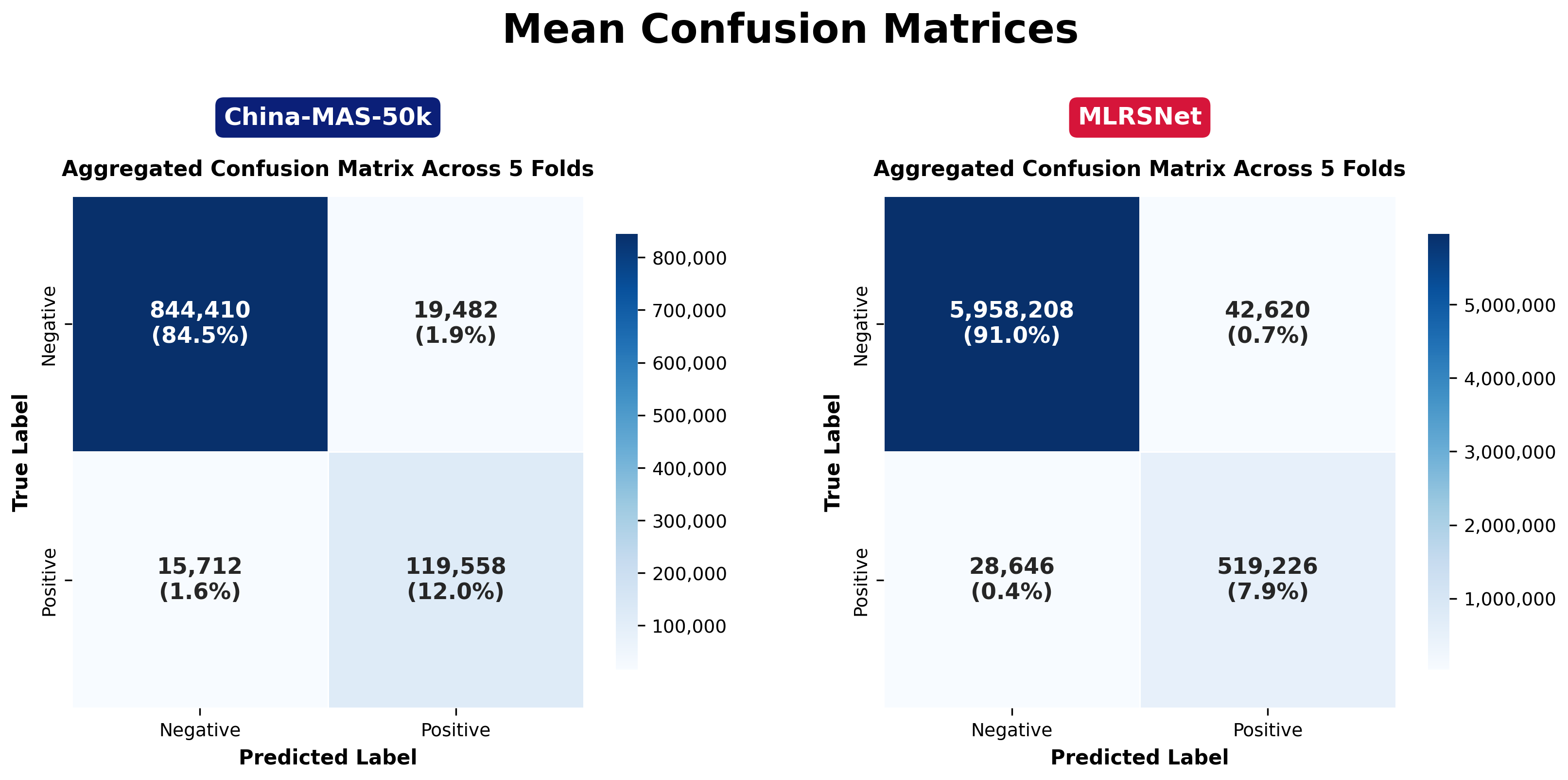}
\caption{Mean Confusion Matrix of the Proposed Method on the Considered Datasets.}
\label{fig:confusion}
\end{figure*}

\subsection{Receiver Operating Characteristic}
The micro-averaged ROC curves of SRCWDNet for the two benchmark datasets are shown in Fig.~\ref{fig:roc}. The tight inter-fold variance (fold AUCs 0.992--0.993 and 0.974--0.978, respectively) further shows the stable discriminative performance across independent test splits, and the relatively balanced label distribution of MLRSNet compared to the long-tailed China-MAS-50k.

\subsection{Precision--Recall Curves}
Micro-averaged precision--recall curves are shown in Fig.~\ref{fig:pr} and are more informative than ROC in the presence of class imbalance. SRCWDNet shows a high mean average precision (AP) of 0.958 on MLRSNet and 0.918 on China-MAS-50k; both datasets maintain precision above 0.9 when recall is not too high, but precision drops as soon as recall goes beyond around 80\%, indicating the challenge of correctly identifying the rare, long-tailed classes at high recall rates.

\subsection{Aggregated Error Analysis}
Fig.~\ref{fig:error} shows the top-15 per-class error rates aggregated over five folds, for the categories that are difficult to classify. The highest error rates are found for the classes Woodland, Road, Bare Land and Cropland, similar to those found for the neighboring land cover classes on China-MAS-50k, while on MLRSNet the most error-prone class is ``bare soil'' ($\sim$11\%), followed by grass and trees, showing that fine-grained natural-terrain categories are still the main contributors to the residual errors on both datasets.

\subsection{Aggregated Confidence Analysis}
Fig.~\ref{fig:confidence} shows the model confidence distribution ($|2p-1|$) for correct and incorrect predictions, averaged over all folds and classes. The model shows good calibration overall, with a large number of predictions falling in a small number of confidence bands for both datasets (the vast majority of low to moderate confidence), and the failure cases being dispersed above the other bands of confidence, with a statistically significant number of predictions at high confidence in China-MAS-50k but not in MLRSNet.

\subsection{Mean Confusion Matrix}
The confusion matrices (all classes, all folds) for each dataset are shown as aggregated $2\times2$ confusion matrices in Fig.~\ref{fig:confusion}. Overall, 91.0\% of all label decisions are correct on MLRSNet, and 7.9\% are correct on China-MAS-50k, with a comparatively higher mistake rate of 1.1\% (false negatives + false positives) and 3.5\% on the latter dataset due to the higher class imbalance and long-tailed label distribution.

\section{Conclusion}
In this paper, we proposed a novel framework for multi-label remote sensing scene classification, namely SRCWDNet, which combines Cross-Window Attention, Dual Label-Graph reasoning and a Neural Classifier Chain into a trainable pipeline. Through extensive experiments on MLRSNet and China-MAS-50k, the results are consistently strong (98.93\% and 96.53\% accuracy, respectively), surpassing the existing CNN- and transformer-based baselines and explicitly modeling label co-occurrence and sequential dependencies that were largely ignored by prior methods. The larger difference seen in the performance on the China-MAS-50k test sets demonstrates the advantage of adapting to class imbalance in the presence of related labels. Future work will involve a lightweight backbone for more computational efficiency, extension to other multi-label RS benchmarks, and calibration-aware relational reasoning with noisy or missing label supervision.



\begin{thebibliography}{48}
\bibitem{b1} M. Drusch \textit{et al.}, ``Sentinel-2: ESA's optical high-resolution mission for GMES operational services,'' \textit{Remote Sens. Environ.}, vol. 120, pp. 25--36, 2012.
\bibitem{b2} X. Qi \textit{et al.}, ``MLRSNet: A multi-label high spatial resolution remote sensing dataset for semantic scene understanding,'' \textit{ISPRS J. Photogramm. Remote Sens.}, vol. 169, pp. 337--350, 2020.
\bibitem{b3} M. R. Boutell, J. Luo, X. Shen, and C. M. Brown, ``Learning multi-label scene classification,'' \textit{Pattern Recognit.}, vol. 37, no. 9, pp. 1757--1771, 2004.
\bibitem{b4} G. B{\"u}ttner, J. Feranec, G. Jaffrain, L. Mari, G. Maucha, and T. Soukup, ``The CORINE land cover 2000 project,'' \textit{EARSeL eProc.}, vol. 3, no. 3, pp. 331--346, 2004.
\bibitem{b5} X. X. Zhu \textit{et al.}, ``Deep learning in remote sensing: A comprehensive review and list of resources,'' \textit{IEEE Geosci. Remote Sens. Mag.}, vol. 5, no. 4, pp. 8--36, 2017.
\bibitem{b6} C. Cortes and V. Vapnik, ``Support-vector networks,'' \textit{Mach. Learn.}, vol. 20, no. 3, pp. 273--297, 1995.
\bibitem{b7} L. Breiman, ``Random forests,'' \textit{Mach. Learn.}, vol. 45, no. 1, pp. 5--32, 2001.
\bibitem{b8} M.-L. Zhang and Z.-H. Zhou, ``ML-KNN: A lazy learning approach to multi-label learning,'' \textit{Pattern Recognit.}, vol. 40, no. 7, pp. 2038--2048, 2007.
\bibitem{b9} J. Read, B. Pfahringer, G. Holmes, and E. Frank, ``Classifier chains for multi-label classification,'' \textit{Mach. Learn.}, vol. 85, no. 3, pp. 333--359, 2011.
\bibitem{b10} S. Koda, A. Zeggada, F. Melgani, and R. Nishii, ``Spatial and structured SVM for multilabel image classification,'' \textit{IEEE Trans. Geosci. Remote Sens.}, vol. 56, no. 10, pp. 5948--5960, 2018.
\bibitem{b11} M.-L. Zhang and Z.-H. Zhou, ``A review on multi-label learning algorithms,'' \textit{IEEE Trans. Knowl. Data Eng.}, vol. 26, no. 8, pp. 1819--1837, 2013.
\bibitem{b12} L. Zhang, L. Zhang, and B. Du, ``Deep learning for remote sensing data: A technical tutorial on the state of the art,'' \textit{IEEE Geosci. Remote Sens. Mag.}, vol. 4, no. 2, pp. 22--40, 2016.
\bibitem{b13} A. Zeggada, F. Melgani, and Y. Bazi, ``A deep learning approach to UAV image multilabeling,'' \textit{IEEE Geosci. Remote Sens. Lett.}, vol. 14, no. 5, pp. 694--698, 2017.
\bibitem{b14} G. Sumbul and B. Dem{\.I}r, ``A deep multi-attention driven approach for multi-label remote sensing image classification,'' \textit{IEEE Access}, vol. 8, pp. 95934--95946, 2020.
\bibitem{b15} R. Stivaktakis, G. Tsagkatakis, and P. Tsakalides, ``Deep learning for multilabel land cover scene categorization using data augmentation,'' \textit{IEEE Geosci. Remote Sens. Lett.}, vol. 16, no. 7, pp. 1031--1035, 2019.
\bibitem{b16} Y. Hua, L. Mou, and X. X. Zhu, ``Relation network for multilabel aerial image classification,'' \textit{IEEE Trans. Geosci. Remote Sens.}, vol. 58, no. 7, pp. 4558--4572, 2020.
\bibitem{b17} B. Chaudhuri, B. Demir, S. Chaudhuri, and L. Bruzzone, ``Multilabel remote sensing image retrieval using a semisupervised graph-theoretic method,'' \textit{IEEE Trans. Geosci. Remote Sens.}, vol. 56, no. 2, pp. 1144--1158, 2017.
\bibitem{b18} Y. Li, R. Chen, Y. Zhang, M. Zhang, and L. Chen, ``Multi-label remote sensing image scene classification by combining a convolutional neural network and a graph neural network,'' \textit{Remote Sens.}, vol. 12, no. 23, p. 4003, 2020.
\bibitem{b19} J. Kang, R. Fernandez-Beltran, D. Hong, J. Chanussot, and A. Plaza, ``Graph relation network: Modeling relations between scenes for multilabel remote-sensing image classification and retrieval,'' \textit{IEEE Trans. Geosci. Remote Sens.}, vol. 59, no. 5, pp. 4355--4369, 2020.
\bibitem{b20} B. Ma \textit{et al.}, ``Label-driven graph convolutional network for multilabel remote sensing image classification,'' \textit{IEEE J. Sel. Topics Appl. Earth Observ. Remote Sens.}, vol. 17, pp. 2245--2255, 2023.
\bibitem{b21} X. Wang, L. Duan, and C. Ning, ``Global context-based multilevel feature fusion networks for multilabel remote sensing image scene classification,'' \textit{IEEE J. Sel. Topics Appl. Earth Observ. Remote Sens.}, vol. 14, pp. 11179--11196, 2021.
\bibitem{b22} Y. Bazi, L. Bashmal, M. M. A. Rahhal, R. A. Dayil, and N. A. Ajlan, ``Vision transformers for remote sensing image classification,'' \textit{Remote Sens.}, vol. 13, no. 3, p. 516, 2021.
\bibitem{b23} X. Tan, Z. Xiao, J. Zhu, Q. Wan, K. Wang, and D. Li, ``Transformer-driven semantic relation inference for multilabel classification of high-resolution remote sensing images,'' \textit{IEEE J. Sel. Topics Appl. Earth Observ. Remote Sens.}, vol. 15, pp. 1884--1901, 2022.
\bibitem{b24} J. Ji, W. Jing, G. Chen, J. Lin, and H. Song, ``Multi-label remote sensing image classification with latent semantic dependencies,'' \textit{Remote Sens.}, vol. 12, no. 7, p. 1110, 2020.
\bibitem{b25} C. Liu, G. Perantoni, L. Bruzzone, and X. X. Zhu, ``Adaptive gradient calibration for single-positive multi-label learning in remote sensing image scene classification,'' \textit{IEEE Trans. Geosci. Remote Sens.}, 2026.
\bibitem{b26} A. Sharifi, M. M. Safari, and B. Alabdullah, ``Advancing coastal land use mapping through deep multi-label classification and multi-sensor data fusion,'' \textit{Egypt. J. Remote Sens. Space Sci.}, vol. 28, no. 4, pp. 681--689, 2025.
\bibitem{b27} R. Du, X. Tang, J. Ma, X. Zhang, and L. Jiao, ``MLMamba: A Mamba-based efficient network for multi-label remote sensing scene classification,'' \textit{IEEE Trans. Circuits Syst. Video Technol.}, vol. 35, no. 7, pp. 6245--6258, 2025.
\bibitem{b28} M. Stoimchev, B. Koloski, J. Levati{\'c}, D. Kocev, and S. D{\v{z}}eroski, ``HELM: Hierarchical and explicit label modeling with graph learning for multi-label image classification,'' arXiv:2603.11783, 2026.
\bibitem{b29} B. Koloski, M. Stoimchev, J. Levati{\'c}, D. Kocev, and S. D{\v{z}}eroski, ``MAPLE: Multi-path adaptive propagation with level-aware embeddings for hierarchical multi-label image classification,'' arXiv:2603.29784, 2026.
\bibitem{b30} R. Du, X. Tang, J. Ma, and X. Zhang, ``Multi-label remote sensing image classification via scene-to-label learning,'' in \textit{Proc. IEEE Int. Geosci. Remote Sens. Symp. (IGARSS)}, 2025, pp. 7288--7292.
\bibitem{b31} A. Mudraje, B. B. Moser, S. Frolov, and A. Dengel, ``Multi-label scene classification in remote sensing benefits from image super-resolution,'' in \textit{Proc. IEEE Int. Geosci. Remote Sens. Symp. (IGARSS)}, 2025, pp. 8435--8440.
\bibitem{b32} M. Ragab, ``Multi-label scene classification on remote sensing imagery using modified Dingo optimizer with deep learning,'' \textit{IEEE Access}, vol. 12, pp. 11879--11886, 2023.
\bibitem{b33} R. Du, X. Tang, J. Ma, X. Zhang, F. Liu, and L. Jiao, ``Semantic-assisted feature integration network for multilabel remote sensing scene classification,'' \textit{IEEE Trans. Geosci. Remote Sens.}, vol. 63, pp. 1--15, 2024.
\bibitem{b34} H. You, J. Gu, and W. Jing, ``Multi-label remote sensing image land cover classification based on a multi-dimensional attention mechanism,'' \textit{Remote Sens.}, vol. 15, no. 20, p. 4979, 2023.
\bibitem{b35} L. Ma, Y. Zhang, M. Yu, and Y. Pi, ``DARC: Lightweight density-adaptive label relation calibration for multi-label remote sensing scene classification,'' \textit{Appl. Sci.}, vol. 16, no. 15, p. 7681, 2026.
\bibitem{b36} J. Lv and D. Wu, ``SDRCNet: A lightweight structure-guided dual-relation consensus network for optical remote sensing images,'' \textit{Remote Sens.}, vol. 18, no. 15, p. 2565, 2026.
\bibitem{b37} I. Dimitrovski, I. Kitanovski, D. Kocev, and N. Simidjievski, ``Current trends in deep learning for Earth observation: An open-source benchmark arena for image classification,'' \textit{ISPRS J. Photogramm. Remote Sens.}, vol. 197, pp. 18--35, 2023.
\bibitem{b38} S. Yuan \textit{et al.}, ``A multi-label dataset for China's agricultural and rural scenes classification from VHR satellite imagery,'' \textit{Sci. Data}, vol. 13, no. 1, p. 384, 2026.
\bibitem{b39} A. Krizhevsky, I. Sutskever, and G. E. Hinton, ``ImageNet classification with deep convolutional neural networks,'' \textit{Commun. ACM}, vol. 60, no. 6, pp. 84--90, 2017.
\bibitem{b40} K. Simonyan and A. Zisserman, ``Very deep convolutional networks for large-scale image recognition,'' arXiv:1409.1556, 2014.
\bibitem{b41} M. Tan and Q. Le, ``EfficientNet: Rethinking model scaling for convolutional neural networks,'' in \textit{Proc. Int. Conf. Mach. Learn. (ICML)}, 2019, pp. 6105--6114.
\bibitem{b42} Z. Liu, H. Mao, C.-Y. Wu, C. Feichtenhofer, T. Darrell, and S. Xie, ``A ConvNet for the 2020s,'' in \textit{Proc. IEEE/CVF Conf. Comput. Vis. Pattern Recognit. (CVPR)}, 2022, pp. 11966--11976.
\bibitem{b43} K. He, X. Zhang, S. Ren, and J. Sun, ``Deep residual learning for image recognition,'' in \textit{Proc. IEEE Conf. Comput. Vis. Pattern Recognit. (CVPR)}, 2016, pp. 770--778.
\bibitem{b44} Z. Liu \textit{et al.}, ``Swin transformer: Hierarchical vision transformer using shifted windows,'' in \textit{Proc. IEEE/CVF Int. Conf. Comput. Vis. (ICCV)}, 2021, pp. 9992--10002.
\bibitem{b45} T. Chen and C. Guestrin, ``XGBoost: A scalable tree boosting system,'' in \textit{Proc. 22nd ACM SIGKDD Int. Conf. Knowl. Discov. Data Min.}, 2016, pp. 785--794.
\bibitem{b46} C. Szegedy, V. Vanhoucke, S. Ioffe, J. Shlens, and Z. Wojna, ``Rethinking the inception architecture for computer vision,'' in \textit{Proc. IEEE Conf. Comput. Vis. Pattern Recognit. (CVPR)}, 2016, pp. 2818--2826.
\bibitem{b47} S. Xie, R. Girshick, P. Doll{\'a}r, Z. Tu, and K. He, ``Aggregated residual transformations for deep neural networks,'' in \textit{Proc. IEEE Conf. Comput. Vis. Pattern Recognit. (CVPR)}, 2017, pp. 5987--5995.
\bibitem{b48} G. Huang, Z. Liu, L. van der Maaten, and K. Q. Weinberger, ``Densely connected convolutional networks,'' in \textit{Proc. IEEE Conf. Comput. Vis. Pattern Recognit. (CVPR)}, 2017, pp. 2261--2269.
\end{thebibliography}
\end{document}